\documentclass{article}

\usepackage{PRIMEarxiv}

\usepackage{lineno}
\usepackage{natbib}

\usepackage[utf8]{inputenc} % allow utf-8 input
\usepackage[T1]{fontenc}    % use 8-bit T1 fonts
\usepackage{hyperref}       % hyperlinks
\usepackage{url}            % simple URL typesetting
\usepackage{booktabs}       % professional-quality tables
\usepackage{amsfonts}       % blackboard math symbols
\usepackage{nicefrac}       % compact symbols for 1/2, etc.
\usepackage{microtype}      % microtypography
\usepackage{fancyhdr}       % header
\usepackage{graphicx}       % graphics
\graphicspath{{media/}}     % organize your images and other figures under media/ folder

\usepackage{amsmath}
\usepackage{multirow}
\usepackage{algorithm}
\usepackage{algpseudocode}
\usepackage{pdflscape}
\usepackage{xcolor}

\DeclareMathOperator*{\argmax}{arg\,max}

\title{A Universal Adversarial Policy for Text Classifiers
\thanks{\textit{\underline{Citation}}: 
\textbf{G. Maimon and L. Rokach, A universal adversarial policy for text classifiers. Neural Networks (2022), \href{https://doi.org/10.1016/j.neunet.2022.06.018}{https://doi.org/10.1016/j.neunet.2022.06.018}}}}

\author{
  Gallil Maimon, Lior Rokach \\
  Ben-Gurion University of the Negev \\
  Beer Sheva\\
  \texttt{\{gallilm, liorrk\}@post.bgu.ac.il} \\
}

\begin{document}
\maketitle

\begin{abstract}
Discovering the existence of universal adversarial perturbations had large theoretical and practical impacts on the field of adversarial learning. In the text domain, most universal studies focused on adversarial prefixes which are added to all texts. However, unlike the vision domain, adding the same perturbation to different inputs results in noticeably unnatural inputs. Therefore, we introduce a new universal adversarial setup - a universal adversarial policy, which has many advantages of other universal attacks but also results in valid texts - thus making it relevant in practice. We achieve this by learning a single search policy over a predefined set of semantics preserving text alterations, on many texts. This formulation is universal in that the policy is successful in finding adversarial examples on new texts efficiently. Our approach uses text perturbations which were extensively shown to produce natural attacks in the non-universal setup (specific synonym replacements). We suggest a strong baseline approach for this formulation which uses reinforcement learning. It's ability to generalise (from as few as 500 training texts) shows that universal adversarial patterns exist in the text domain as well.
\end{abstract}

% keywords can be removed

\section{Introduction and Motivation} 
\label{sec:intro}
Leading deep learning models have been shown to be sensitive to adversarial attacks. These are small input perturbations which induce wrong predictions. \citet{moosavi2017universal} showed the existence of Universal Adversarial Perturbations (UAP), which are input independent perturbations which induce wrong predictions on many inputs. This helped develop current leading mental models of adversarial learning (``Adversarial Examples are not Bugs they are Features'')\citep{ilyas2019adversarial} thus deepened our understanding of adversarial examples. The ability of such perturbations to generalise to unseen inputs also has practical benefits regarding model access and efficiency when performing attacks on many new texts.

\begin{figure*}[!ht]
  \centering
  \includegraphics[width=\linewidth]{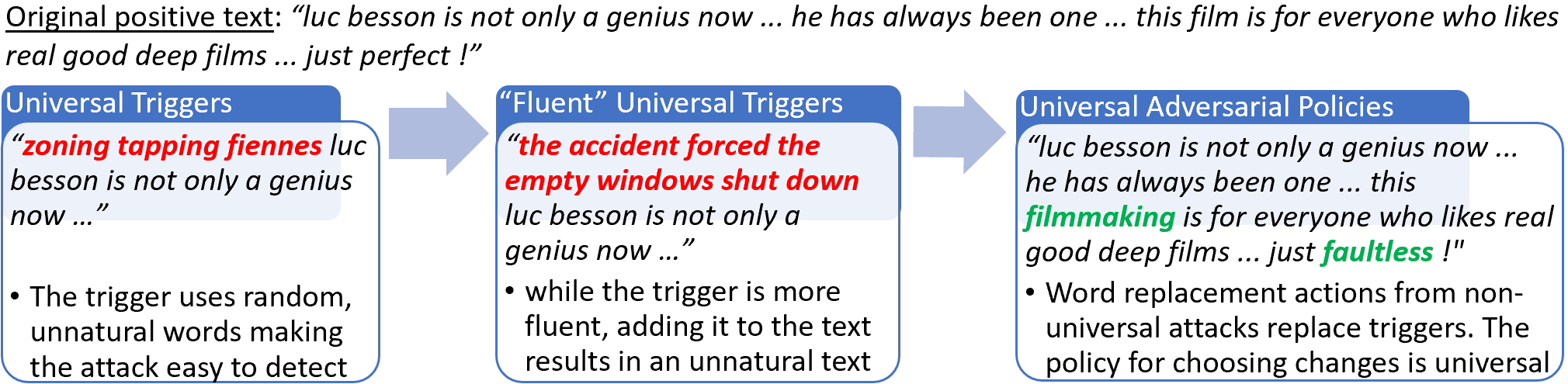}
  \caption{Advances in text universal adversarial approaches, from unconstrained trigger attacks, to attempts to create fluent triggers. As the example shows, even ``fluent" triggers can often result in unnatural texts. Therefore, we suggest using synonym perturbations in universal settings as well. Trigger examples are from respective papers, while the synonym attack was generated using LUNATC.}
  \label{fig:universal_attacks}
\end{figure*}

In the text domain, most studies on universal adversarial attacks focused on adding a single perturbation (akin to other domains), in this case - an adversarial prefix \citep{behjati2019universal}. While the reported fooling rates were high, the sequence added was often nonsensical and was clearly noticeable to humans due to the unnatural and ungrammatical resulting texts. This limitation was addressed by \Citet{song2020universal}, who suggested a method for generating more fluent adversarial prefixes. While this improved the naturalness of the triggers, adding the same prefix is still limited and often leads to noticeably unnatural inputs (see Figure \ref{fig:universal_attacks}).

\textcolor{black}{In the non-universal setup, leading methods for generating adversarial examples to text classifiers, focus on finding such examples in a search space that has been predefined by a set of semantically preserving alterations. Many experiments were conducted (including human evaluation) which show these approaches create adversarial examples which seem natural and preserve semantics, when the search space is defined properly}  \citep{morris-etal-2020-reevaluating}. For this reason, they are preferred to attack methods which alter the text in other ways (e.g. add texts). Search-based attacks weren't used in universal settings because others claimed ``word-replacing and embedding-perturbing approaches ... (are) not applicable" \citep{song2020universal}.

\begin{figure*}[!ht]
  \centering
  \includegraphics[scale=0.55]{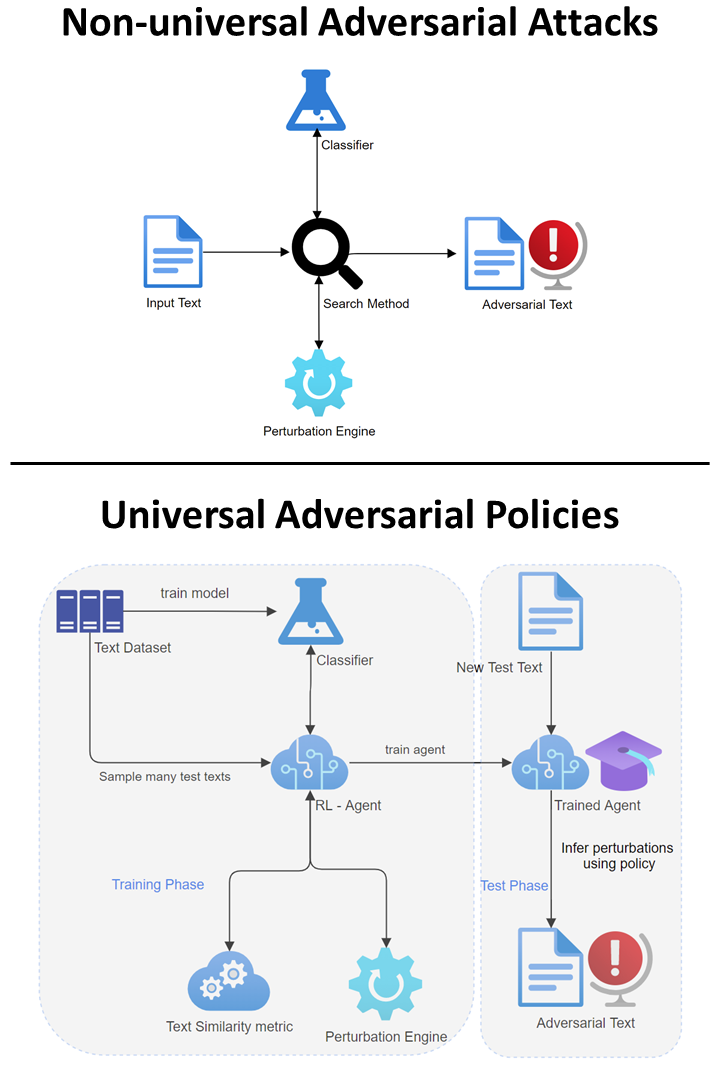}
  \caption{\textcolor{black}{A comparison between the universal adversarial policy setup and non-universal attacks. Universal approaches have a training phase in which they ``learn" from experience in adversarially attacking many texts from the same domain. Then, they efficiently attack new texts in the test phase. On the other hand, non-universal approaches attack each text individually without using previous texts. Instead they use extensive or heuristic search methods to find adversarial texts.}}
  \label{fig:universal_attack_setup}
\end{figure*}

We introduce a new form of a universal adversarial formulation - \textbf{universal adversarial policy}, which is the first universal approach to use such word-replacing methods, thus resulting in relevant, natural texts. Instead of generating a single perturbation, one learns a single, parametric search policy, over a predefined set of text perturbations. Like the non-universal approach the search objective is to find a text which changes the prediction of the attacked model, but is as similar to the original as possible. However, in this setup, the search policy is learned and parameterised so that it can generalise to new, unseen texts. \textcolor{black}{An overview of this approach is shown in figure \ref{fig:universal_attack_setup}}. We evaluate the policy's ability to find adversarial examples on a set of unseen texts, with a single action ``ordering" as further explained in section \ref{sec:problem_formulation}.

\textcolor{black}{While this does not create universal \textbf{perturbations}, because the perturbations are dependant on the input, the universality of the \textbf{policy} is still interesting. This formulation maintains many practical benefits of UAPs because it makes generating attacks for unseen texts more efficient regarding oracle access and run time. For instance, if we wanted to generate many toxic comments on Wikipedia which wouldn't be detected as such. Non-universal methods would attack each comment separately, whereas universal policies, will utilise the experience from previous comments to efficiently attack new comments. In the future, we hope that universal adversarial perturbations will alleviate the need for test time model access altogether as discussed in section \ref{sec:conclusion}. In addition, should a universal policy succeed in generalising to unseen inputs, it would indicate that universal adversarial patterns exist in the text domain as well, which means that the direction of adversarial examples is not independent on the input. This will hopefully help advance our theoretical understanding in textual adversarial learning.} Finally, The existence of global patterns to adversarial attacks can help better understand specific model biases, and the learning process of models in general. While previous universal attacks in text domain existed, we find this framework is more suitable for text data as it is inline with per-text (non-universal) adversarial perturbation research. This allows it to benefit from guarantees about the naturalness and the semantics of the adversarial texts. We suggest a reinforcement learning (RL) based method for \textbf{l}earning a \textbf{un}iversal \textbf{a}dversarial policy for \textbf{t}ext \textbf{c}lassifiers (LUNATC), as a strong baseline for this formulation.

In this study, our main contributions are:
\begin{itemize}
\item Describing a new formulation for universal adversarial attacks - \textit{universal adversarial policies}, which is well-suited to the unique properties of text.
\item Introducing LUNATC which is a novel, model-agnostic, black-box algorithm for learning a universal adversarial policy to text classifiers, using Deep Q-learning \citep{mnih2013playing}, and publishing the code\footnote{\href{https://github.com/gallilmaimon/LUNATC}{https://github.com/gallilmaimon/LUNATC}}.
\item Publishing a classification dataset, based on Pubmed papers\footnote{Courtesy of the U.S. National Library of Medicine}. It has substantially more samples than other common datasets, thus we hope it will help further the research of generalisation in text adversarial examples.
\end{itemize}

\section{Related Work}
\subsection{Search Based Attacks}
These approaches generate a different perturbation for each input text, by searching a predefined search space of text alterations for the best attack. The perturbations aim to be semantics preserving and natural, and most commonly work at word-level such as synonym replacement \citep{jin2020bert} or named entity replacement \citep{ren-etal-2019-generating}. Other methods also used character-level changes, such as misspelling \citep{li2018textbugger}, and some used machine re-translation to offer more global changes \citep{ribeiro-etal-2018-semantically}. Examples of attacks created using synonym replacement can be seen in Table \ref{attack-examples}. Each attack uses a different search method, from greedy heuristics \citep{ren-etal-2019-generating, jin2020bert} to more computationally intensive methods \citep{alzantot-etal-2018-generating}. A recent survey suggested decomposing the definition of the search space from the effectiveness of the search algorithm \citep{morris-etal-2020-reevaluating}. They also showed that using the synonym search space, constrained with high similarity thresholds, results in high quality attacks.

\textcolor{black}{These search based approaches can be thought of as \textbf{non-universal} policies. This means that the search policy does not use ``experience" from successfully attacking other texts in the domain to efficiently attack new texts. Instead, they attack each text individually and decide which perturbations to perform based on access to the attacked model's predictions.}

\begin{table}[t]
\caption{Successful attack examples. Original text and correct, predicted label appear first, then the synonym-based attack which caused the model to mis-classify the example.}
\label{attack-examples}
\centering
\begin{tabular}{ll}
\toprule
 \multicolumn{2}{c}{Toxic Wikipedia comment detection (Toxic-Not)} \\
\midrule
Not     & = ughh ....= my god , middletown is a horrible place ! delaware \\ & needs to pass some anti - incest laws ... \\
Attack & = ughh ....= my god , middletown is a  \textbf{terrible} place ! delaware \\ &  needs to pass some anti - incest laws ... \\
\midrule
\multicolumn{2}{c}{IMDB reviews (Positive-Negative)} \\
\midrule
Pos    & luc besson is not only a genius now ... he has always been one ...\\ &  this film is for everyone who likes real good deep films ... \\ & just perfect ! \\
Attack & luc besson is not only a genius now ... he has always been one ...\\ & this \textbf{filmmaking} is for everyone who likes real good deep films ... \\ & just \textbf{faultless} ! \\
\midrule
\multicolumn{2}{c}{Pubmed abstract (Review-Case study)} \\
\midrule
Review & to report 11 cases of possible erythromycin - induced hearing loss\\ & and to review all cases reported in the literature . \\
Attack & to report 11 \textbf{case} of possible erythromycin - induced hearing loss \\ & and to review all cases reported in the literature .  \\
\bottomrule
  \end{tabular}
\end{table}

\subsection{Universal Trigger Attacks}
\Citet{behjati2019universal} and \citet{wallace2019universal} suggest universal adversarial triggers as a form of UAP for text. This involves using gradient projection in word embedding space to find the sequence most likely to alter the prediction when added to the beginning of texts. This approach had high success rates in altering the prediction with few words, but, the resulting texts were highly unnatural and could be easily detected by humans (see Figure \ref{fig:universal_attacks}).  \citet{song2020universal} tried to address this, by adding an adversarially regularised auto-encoder which aims to enforce the naturalness of the generated prefix. However, as mentioned in Section \ref{sec:intro}, and demonstrated in Figure \ref{fig:universal_attacks}, using a single trigger, even if ``fluent", on many texts inevitably leads to many unnatural texts.

\section{Problem Formulation} \label{sec:problem_formulation}
We introduce a new formulation for adversarial attacks which has many benefits of universal attacks, while maintaining the quality guarantee of per-sample attacks. We do this by making word-replacing attacks applicable in a universal context. This approach suggests learning a universal adversarial \textit{policy}. A universal adversarial policy is a learnable search policy, over a predefined search space of semantics preserving perturbations. The search objective is finding a text which is differently classified by the attacked classifier, while being as similar to the original text and as natural as possible. Our search policy is learned on a set of texts but is required to be \textit{universal} in that the policy performs well on unseen texts (from the same task and domain) as well. 

Formally, we have a set of training texts $\mathcal{S}_{tr}$, an unknown set of test texts $\mathcal{S}_{te}$, \textcolor{black}{and} an attacked classifier $\mathcal{C}$\textcolor{black}{.} \textcolor{black}{The set of text perturbations, is defined by the transition function $\delta$, defined as: $\delta(t, i) = t'$, where $t$ is an input text, the index $i$ indicates different perturbations and $t'$ is the perturbed text.} The different perturbations are defined by an index $i$ and can be at word-level (like synonym-replacement), at character level (such as misspelling) or at text level (e.g. machine re-translation). We will use synonym replacement as the only perturbation for simplicity of the explanation in this running example, but the generality isn't limited in any way. In this case, the action $i$ indicates replacing the word at the $i$'th location in text $t$ with a suitable synonym. \textcolor{black}{Our specific approach for selecting appropriate synonyms is discussed in section \ref{subsec:action}.}

\textcolor{black}{Intuitively, a policy $\mathrm{P}$, should receive an input text $t$ and output the actions needed to reach a perturbed text which is the highest quality adversarial attack. For instance, given the input text \textit{``I loved this movie"}, we would want the policy to suggest replacing \textit{``loved"} and then \textit{``movie"} with synonyms resulting in \textit{``I liked this film"} which is hopefully misclassified by the attacked classifier. This makes it an adversarial example. It is worth noting that the choice of the synonym to replace a given word is dependant on the current text (as further explained in section \ref{subsec:action}, therefore the order of actions is important. As explained, actions are defined by indices. Thus, formally, a} policy $\mathrm{P}$ parameterised by $\theta$, defines an ordering of all possible actions for any input text:

\begin{equation}
\label{eq:policy_definition}
    \mathrm{P}_\theta(t) = j_0, j_1, j_2, ... \quad, s.t. \quad j_i \in \mathcal{N}_{actions}
\end{equation}

\noindent This policy essentially defines a specific search path from the initial \textcolor{black}{text} and does not traverse the entire search space. The policy can be used to receive several different search paths from the initial state (if it is a statistical model), thus covering more of the search space. In this setup we focus on the single, best search path, because we are interested in efficient inference on new texts.

This ordering \textcolor{black}{ of the actions, outputted by the policy} defines a series of texts, $t'_i$, defined by the following recursive rule:

\begin{equation}
    t'_i = \\
    \begin{cases}
    t, \qquad \qquad \qquad if \quad i=0\\
    \delta(t'_{i-1}, \mathrm{P}_\theta(t)_{i-1}) \qquad else \\
    \end{cases}
\end{equation}

\noindent This series of texts starts at the initial text $t$, followed by the text achieved by applying the first action outputted by the policy on the text. Then we apply the second action on the new text and so on.

For instance, given that we use synonym replacement as the only action, then the text \textit{``I loved this movie"}, would have 4 possible actions (matching the 4 words which can be replaced), and a policy could output this ordering of the actions: $\mathrm{P}_\theta(t) = 3, 1, 2, 0$. This would define the text series: starting with the original text - \textit{``I loved this movie"}, then the text after performing the first action which is replacing the word at index \textit{3} resulting in $\rightarrow$ \textit{``I loved this film"}. We then perform the second action which the policy outputted, on our previous text, replacing the word at index \textit{1} resulting in $\rightarrow$ \textit{``I liked this film"} and so on.

Based on the outputted ordering, we define the adversarial text as the first perturbation which changes the classification of the attacked model $\mathcal{C}$, if such one exists. More formally, the final text $\mathcal{T}$ is defined as follows:

\begin{equation}
    \mathcal{T}(\mathrm{P}_\theta, \mathcal{C}, t) = t'_i \ for \ minimal \ i \ s.t. \ \mathcal{C}(t'_i)\neq\mathcal{C}(t)
\end{equation}

\noindent For instance in our running example - if our sentiment classifier correctly classified \textit{``I loved this movie"} and \textit{``I loved this film"} as positive, but classified \textit{``I liked this film"} as negative - then \textit{``I liked this film"} would be the adversarial text. 

Scoring metrics for adversarial texts can vary to consider the prediction change, the naturalness and the similarity to the original text. For simplicity, we use a scoring function which receives the semantic similarity score defined by the Universal Sentence Encoder (USE) \citep{cer-etal-2018-universal} if an attack exists, and 0 otherwise. \textcolor{black}{This method for semantic similarity is the standard practice, which was first shown to correlate to human labels in the original paper. This method also became the the standard for evaluating the quality of text adversarial attacks in Textfooler and was shown by \citet{morris2020second} to be effective compared to other methods such as BERTScore.} We mark this score as $S(\mathrm{P}_\theta, t)$.

\begin{equation}
    S(\mathrm{P}_\theta, t) = semantic\_sim(t, \mathcal{T}(\mathrm{P}_\theta, \mathcal{C}, t))
\end{equation}

So in our running example $S(\mathrm{P}_\theta, t)$ = semantic\_sim(\textit{``I loved this movie"}, \textit{``I liked this film"}), i.e - the semantic similarity of the adversarial text to the original text. Semantic\_sim is the cosine similarity of the embeddings of the texts according to USE. If all the texts in the series were classified correctly then the score would be 0. 

An optimal universal policy defines the highest scoring adversarial texts on texts in the test set - $\mathcal{S}_{te}$, \textbf{by optimising $\theta$ using texts from the training set only}. Formally:

\begin{equation}
\label{eq:optimal_policy}
    \mathrm{P}_{\theta -opt} = \argmax_{\mathrm{P}_\theta}(\mathrm{E}_{t \in \mathcal{S}_{te}}[S(\mathrm{P}_\theta, t)])
\end{equation}

As mentioned, for simplicity, we focus in this paper on a specific search space, defined by a perturbation function of word synonym replacement. Thus $\delta(t, i)$ indicates replacing the word at location $i$ with a synonym. Our approach for selecting appropriate synonyms is explained further in section \ref{subsec:action}. For further simplifying the search space, we do not allow replacing the same word several times, which means that $\mathrm{P}_\theta(t)$ outputs a permutation of all possible changes -  $\sigma(1, \dots, n_{actions})$, and not any ordering with repetitions.

\section{LUNATC Algorithm} \label{sec:our_approach}
\textcolor{black}{The universal adversarial policy can be defined using many parametric search methods, based on classic supervised learning, such as GenFooler introduced in section \ref{subsec:baselines}. However, we believe reinforcement learning is a natural fit for this formulation. As opposed to standard supervised learning, RL, optimises an agent which learns through interacting with an environment. At each observation of the state, the agent must choose which action to perform. For each state and action the agent receives a reward from the environment. The agent tries to maximise the cumulative reward from all the actions it performs.} 

Because it inherently defines a search policy over an action (perturbation) space from an initial text (as a state), it can naturally match the adversarial policy formulation. It also learns state representations which help it generalise to unseen states. Furthermore, RL works well with discrete state and action spaces, and has been shown to be efficient for text manipulation \citep{mirowski2016learning}. In order to solve the task of generating adversarial examples with RL we must define states, actions, rewards and the agent.

\subsection{State}
We wish to start at a given input text, and change it until arriving at a new, altered text, which is hopefully an adversarial example to the attacked model. Therefore, it makes sense to define our states as the texts themselves, with a text we wish to attack being an initial state and a successful adversarial example being a terminal state. Using the general notation from section \ref{sec:problem_formulation}, the initial state would be $t$ and $\mathcal{T}(\mathrm{P}, \mathcal{C}, t)$ would be a terminal state.

Of course, in order to use texts as inputs for the agent, we must represent them as vectors. We follow the text embedding approach suggested in BERT \citep{devlin2018bert} - taking the mean of the last four hidden layers, resulting in a fixed-size representation. We use a pretrained BERT model for this.

\subsection{Action}
\label{subsec:action}
This leads to defining actions as text perturbations which move us between different states. We aim for these actions to preserve the semantic meaning of the text.  This corresponds with the $\delta$ function described in section \ref{sec:problem_formulation}. As stated previously, we use synonym replacement as the only action, and follow the method performed by \Citet{jin2020bert}, while tweaking the similarity thresholds based on \Citet{morris-etal-2020-reevaluating}, thus guaranteeing high-quality attacks. This also helps maintain a fair comparison with other approaches, by decoupling improvements in the search algorithm and changes to the search space (which can introduce non-natural texts), as suggested by \Citet{morris2020textattack}.

\textcolor{black}{More specifically, a list of synonym candidates is suggested using cosine similarity of word vectors specially curated for synonym finding \cite{mrkvsic2016counter}. We take all words above a given similarity threshold. Stop words are then removed using a fixed list. We evaluate the words' part-of-speech within the context, and candidates deemed to have different parts-of-speech are filtered out. Finally, we replace the word with all remaining candidates, and compute the similarity of the resulting texts to the previous texts. Of all texts above a given similarity threshold (according to USE), the one which most changes the attacked model's prediction is selected. If the several replacement options change the model's predicted class, then the most similar text is chosen.}

To aid the generalising abilities of the agent to other texts, we introduce another version of the DQN algorithm which uses an embedded action representation (as further explained in sub-section \ref{subsec:agent}). We aim to induce a prior bias that certain actions are more similar than others, based on the words' meaning and not only their location in the text. To this end we represent words using Glove vectors \citep{pennington-etal-2014-glove}. To represent the action of replacing the word $w$ at location $i$, we use the following formula:

\begin{equation}
\label{eq:action_embed}
    emb\_A(w, i) = word\_vec(w) + \alpha * pos\_enc(i)
\end{equation}
\noindent where pos\_enc is the positional encoding method from BERT, and $\alpha$ a hyper-parameter.

\subsection{Reward}
The reward of a RL task is a crucial part, that needs to balance accurately defining the wanted achievement and learnability of the function. The reward function should correlate to $S(\mathrm{P}, t)$, i.e - a given perturbed text is only good as an adversarial sample if it changes the attacked model's classification compared to the original, with it being considered better the more similar it is to the original. However, leaving the reward as zero for all the non-terminal states poses an exploration problem for the agent. Therefore, we wish to differentiate the reward for intermediate states. These assumptions led us to define the reward function as follows:

\begin{equation}
\small
\label{eq:agent_reward}
    r(S, a)=
    \begin{cases}
      - \varepsilon \qquad \qquad \text{if}\ S' = S \\
      (F_{\widetilde{y}}(S)-F_{!\widetilde{y}}(S)) - max(F_{\widetilde{y}}(S')-F_{!\widetilde{y}}(S'), 0) 
      \\ \qquad \qquad \equiv r_{logit},\qquad \text{if}\ \widetilde{F}(S')=\widetilde{y} \\
      r_{logit} + Semantic\_Sim(S_{init}, S') \qquad\text{else}
    \end{cases}
\end{equation}

\noindent where $F_i$ is the logit of class $i$ by model $F$ and $\widetilde{F}(S)=argmax_i(F_i(S))$ i.e the predicted class. $S'$ is the state reached after performing the action $a$ at state $S$. $S_{init}$ is the initial text. We mark $\widetilde{y}=\widetilde{F}(S_{init})$, \textcolor{black}{which marks the predicted class for the original text and $!\widetilde{y}$ for the next most likely class (in the binary classification case, there is only one class not predicted, in the multi-class case this is the class with the second highest logit)}. Simply put, the reward is a negative constant for actions which make no difference (to deter the agent from making them), and is equal to the the decrease in the gap between the source and target class logits, if the predicted class hasn't changed. If the agent is successful in changing the predicted class, the game ends and the reward is the previous logit reward plus a score of the semantic similarity of the text and the original text. We use cosine similarity of the two texts' embedding using USE. The similarity score is between 0 and 1, but its values are scaled to be between 0 and 100, whereas the logits difference, marked $r_{logit}$, tends to be much lower thus still giving higher weight to the end attack. We add $r_{logit}$ to the similarity reward when the class changes, to maintain that more similar attacks will have higher rewards. This achieves this by making sure that all attacks which change the class will have the same cumulative logit reward (equal to $F_{\widetilde{y}}(S)-F_{!\widetilde{y}}(S)$), regardless of how many the steps they took, and of the confidence gap of the new class. This means that the only difference will have to do with the semantic similarity.

\subsection{Agent}\label{subsec:agent}
Once formulating the task as a RL task, we have a variety of algorithms that can be used \citep{fortunato2017noisy, hessel2018rainbow}. We chose DQN as it is well established and simple. However, we introduce a variation of the standard algorithm which utilises an embedding of the actions as well. While the standard approach approximates $Q(s, a)$ with a neural network which receives $s$ as an input and has $|a|$ outputs, we replace this with a network which receives both $s$ and $a$ as inputs, and has a single output. This variation lets us encode the actions, in a way which utilises our domain understanding to represent actions in a meaningful way. In our case, the use of word vectors, indicates that certain actions are more similar than others. This improves generalisation results as is further studied in the results section. We also find empirically that using target and policy networks, works best in our case, when using the target network for the ``optimal" action selection only. This is slightly different from what is suggested in DQN or double DQN \citep{van2016deep}. However, we find that it better achieves the goal of reducing over estimation and improving the overall results.

Unlike most RL setups we have separate train and test phases. In training we optimise the policy parameters $\theta$ to get closer to the optimal policy defined in eq. \ref{eq:optimal_policy}, using the set of training texts $\mathcal{S}_{tr}$. This phase is described in Algorithm \ref{psuedocode-train}. \textcolor{black}{Lines} 1-3 describe the initialisation of the target and policy network and the memory. lines 4-12 describe the course of each ``round" - a text is sampled from the training set and then actions are performed, based on $\epsilon$-greedy exploration and the policy network. The policy network is updated after each action based on batch of transitions sampled from memory. The round ends when the predicted class has changed, there are no more legal actions or the maximum number of actions is reached. As in DQN we update the target network periodically. In the test phase, we use the trained model to select which perturbations to perform at each stage as in eq. \ref{eq:policy_definition}. This is described in Algorithm \ref{psuedocode-test}

\begin{algorithm}
  \caption{LUNATCTrain($\mathcal{S}_{tr},\mathcal{C}$)}\label{psuedocode-train}
  \hspace*{\algorithmicindent} \textbf{Input:} training texts $\mathcal{S}_{tr}$ and Text classifier $\mathcal{C}$
  \begin{algorithmic}[1]
  \State $n \gets 0, M\gets \{\}$
  \State $\mathcal{A}_{pol}\gets random \ init \ agent$, $\mathcal{A}_{tar}\gets\mathcal{A}_{pol}$ 
  \While{$n<num\_rounds$}
    \State $s \gets \mathcal{S}_{tr}$, $l \gets legal\_actions(s)$, $R \gets 0$
    \While{$\mathcal{C}(s) == \mathcal{C}(s_{init}) \ and \ len(l)>0 \ and \ max\_turns \ not \ reached$}
        \State $emb\_a = emb(s, i)$ for i in $l$ \Comment{eq. \ref{eq:action_embed}}
        \State $a \gets \mathcal{A}(s, emb\_a)$ \Comment{$\epsilon$-greedy}
        \State $s \gets \delta (s,a)$, $R \gets R+ r(s,a)$ \Comment{eq. \ref{eq:agent_reward}}
        \State $l \gets l \setminus a$, $M\gets M \bigcup (s,a,r)$
        \State $b \gets sample \ batch \ from \ M$
        \State $update \ \mathcal{A}_{pol} \ with \ b$  \Comment{by DQN}
    \EndWhile
    \State $n \gets n+1$
    \If{target update round}
    \State $\mathcal{A}_{tar}\gets\mathcal{A}_{pol}$ 
    \EndIf
  \EndWhile
  \State \textbf{return} $\mathcal{A}_{pol}$
  \end{algorithmic}
\end{algorithm}

\begin{algorithm}[t]
  \caption{LUNATCAttack($t \not\in \mathcal{S}_{tr},\mathcal{C}, \mathcal{A}$)}\label{psuedocode-test}
  \hspace*{\algorithmicindent} \textbf{Input:} An unseen text $t$, text classifier $\mathcal{C}$, and policy $\mathcal{A}$
  \begin{algorithmic}[1]
  \State $t_{cur} \gets t$
  \While{$\mathcal{C}(t_{cur}) == \mathcal{C}(t) \ and \ len(l)>0$}
    \State $emb\_a = emb(s, i)$ for i in $l$ \Comment{eq. \ref{eq:action_embed}}
    \State $a \gets \mathcal{A}(s, emb\_a)$ \Comment{no exploration}
    \State $s \gets \delta (s,a)$
    \State $l \gets l \setminus a$
  \EndWhile
  \State \textbf{return} $t_{cur}$
  \end{algorithmic}
\end{algorithm}

\section{Experimental Setup}
All experiments were run on an 8-CPU core, 64 Gb ram machine with one Nvidia-RTX2080 GPU.

\subsection{Datasets}
\textcolor{black}{All datasets used are of text classification. The majority (three) are single text, binary classification tasks on which we focus. However, we also show results on one natural language task with two inout texts and three classes. These are the datasets, relating to various tasks:}

\begin{enumerate}
    \item \textbf{ACL-IMDB} \citep{maas-EtAl:2011:ACL-HLT2011} is a binary sentiment analysis dataset of movie reviews from IMDB.
    \item \textbf{Toxic-Wiki} \cite{Toxic:2017} is a multi-label dataset of comments from Wikipedia, labelled as toxic, obscene, etc. We use the toxic label only for binary classification.
    \item \textbf{Pubmed} is a new dataset we created of binary classification of medical papers' abstracts, that are indexed in the PUBMED search engine, into two categories: review or case report. It has three million samples which is much larger than other existing datasets - enabling research of the influence of the size of $\mathcal{S}_{tr}$ on generalisation.
    \item \textcolor{black}{\textbf{MNLI} \cite{N18-1101} is a dataset of multi-genre natural language inference. This dataset is comprised of pairs of texts, and they are labelled based on} \textcolor{black}{whether the second is derived from the first, contradicts it or is neutral to it.}
\end{enumerate}

All the datasets were pre-processed in the same way for which the code is published. Pre-processing includes removal of HTML tags, and special characters and making the text lower case. Some text examples can be seen in Table \ref{attack-examples}.

In order to focus on the search policy's ability to find attacks and not the definition of the search space, we define $\mathcal{S}_{tr}$ and $\mathcal{S}_{te}$ to only include texts for which an attack is known to exist (in the search space). We do this by attacking the texts with several attacks - Textfooler \citep{jin2020bert}, PWWS \citep{ren-etal-2019-generating}, and a simple search (see section \ref{subsec:baselines}) with 100 rounds, and taking only texts in which any of the approaches were successful. To assess the ability to generalise to attacks found by methods not used to create the training set, $\mathcal{S}_{tr}$ has only texts which Textfooler (TF) was successful on, while $\mathcal{S}_{te}$ also has texts which TF was not. For simplicity of the generalisation all the attacks' ``direction" is the same i.e the correctly predicted, original class of all texts is the same. This means that the agent doesn't need to learn different actions for given states (texts) based on the initial one's class. For IMDB we take positive reviews only, for Wikipedia non-toxic comments and for Pubmed ``review" type abstracts only.

\subsection{Text Classifiers}
We attack two different classifiers to show our method is robust, as in black-box settings the attacked model architecture is unknown. We want to show that using the same architecture (but different weights) for our state representation isn't the reason for success on BERT. One classifier used is a pretrained, base, uncased version of BERT \citep{Wolf2019HuggingFacesTS}. We fine-tune the pretrained model on each dataset, with binary cross-entropy loss, and the Adam optimiser \citep{kingma2014adam}.
Another model is a word-LSTM \cite{hochreiter1997long} model with 1-layer bi-directional LSTM with 150 hidden units, and dropout of 0.3, using 200-dimensional Glove embeddings trained on 6B tokens from Gigawords and Wikipedia. Training code is made available.

\subsection{Compared Algorithms}\label{subsec:baselines}
Our experiments all use the same DQN agent, described previously, with exponentially decaying $\epsilon$-greedy exploration. The agent network architecture is a fully-connected network with six hidden layers. We use the Adam optimiser \citep{kingma2014adam}. We noticed that increasing the number of training texts required increasing the memory size of the agent, the rounds between target network updates, the total number of training steps and the exploration (slower epsilon decay) for better results. This makes sense, because more training texts indicates higher variability in the visited states, thus we need to explore each, and stabilise our learning to update based on more samples. However, due to run-time limitations, the hyper-parameters were not tuned, and standard values were used, somewhat arbitrarily as specified in the code. 
Each round for the agent included sampling a text from $\mathcal{S}_{tr}$, as an initial state. Text selection is ordered so that all texts are sampled before a text is sampled again. In train time, we limit the max number of actions per text to 30 due to run-time considerations, we don't perform this limitation at test time. It doesn't harm results noticeably based on our preliminary study. 

As this is a new formulation, no other baselines exist. We, therefore, introduce two baselines, which do not use RL, and can be seen as generalised versions of TF and PWWS. Both TF and PWWS calculate a word importance value heuristically and use that as an ordering for word replacement. If the perturbation space is only word replacement, this can be generalised to our problem setup. This leads us to introduce Genfooler,  an attack method which tries to learn a mapping from the input texts to the word importance (TF, PWWS or any other). To this end we train a model to predict word importance, given a text. We use the same text embedding method as LUNATC as input for the model, and train it as a multi regressor (with an output for each word in the input text) with a mean squared error loss. We use the Adam optimiser and 5-fold validation to select the optimal number of epochs. We then use the model to predict word importance on the test texts, and perform the attack greedily like PWWS or TF. 

We also introduce a simple search method, which uses the same strong word replacing actions as LUNATC (choosing which word to introduce is done in the same way), however the order of words chosen to be replaced is random. This means that at inference time it has as much access to the attacked model as LUNATC. It also means that any improvement that LUNATC will have over the simple search, is due to it generalising universal adversarial patterns to the new texts.

All of these attacks use the same search space as LUNATC exactly, which is important to isolate the quality of the search method (as shown by \Citet{{morris2020textattack}}). In order to evaluate our approach against methods like BERT-Attack \cite{li2020bert} properly, we would need to implement a version of LUNATC with the same actions which is out of scope for this work.

\section{Results}
To assess attacks, we define the attack \textit{success rate}, as the ratio of successful adversarial examples in $\mathcal{S}_{te}$. An example is successful if it changes the model's classification and its semantic similarity to the original text is above a threshold. We use USE for calculating similarity. The similarity threshold balances quality of attacks and percentage of successful attacks. We specify the threshold used where relevant.

\begin{figure}[t]
  \centering
  \includegraphics[width=\linewidth]{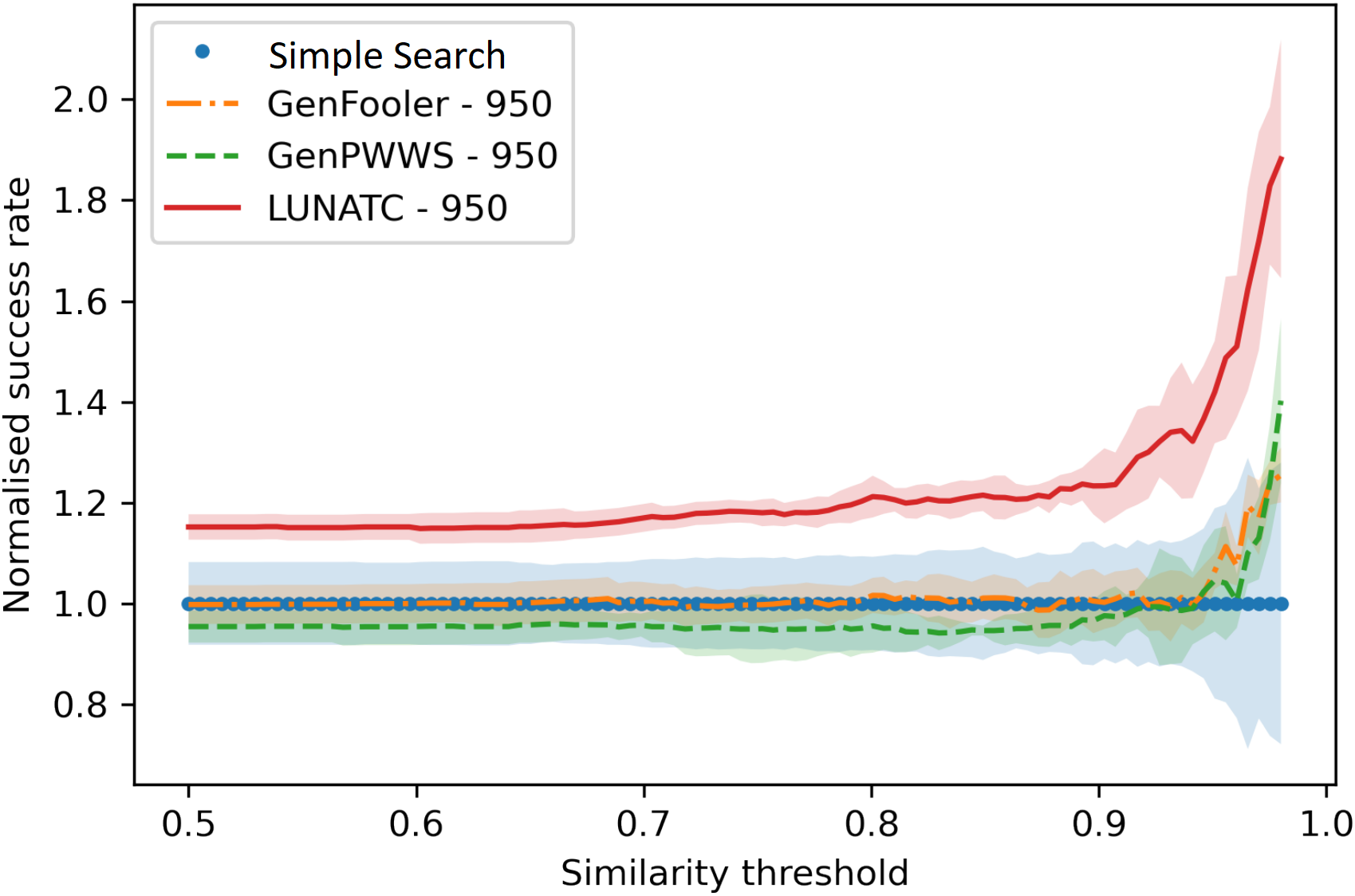}
  \caption{Normalised success rates as a function of the similarity threshold, on the Toxic-Wikipedia dataset, against a BERT classifier. The success rates are normalised by the \textcolor{black}{mean success rate of the simple search approach on the same similarity threshold}. The error margins indicate a 95\% confidence interval across 10 seeds for the simple search approach and three seeds for the rest.}
  \label{fig:sr_similarity}
\end{figure}

To assess the impact of the similarity threshold used as a limit for successful attacks, we plot the normalised success rates of the different universal policies as a function of the threshold in Figure \ref{fig:sr_similarity}. We normalise the success rates by the average success of the simple search method (see section \ref{subsec:baselines}), across 10 seeds. \textcolor{black}{This means that the value describes how much better the results are compared to simple search approach for the same similarity threshold. For instance, we can see in Table \ref{generalised-attack-success} that for a threshold of 0.9, LUNATC\_Max outperforms simple-search on the Toxic-Wikipedia dataset against BERT, by a factor of $\sim$1.24 (37.56 compared to 30.31) which matches what is seen in Figure \ref{fig:sr_similarity}.} These results show that LUNATC outperforms all other attacks across all similarity thresholds. The results also show a greater increase with high similarity thresholds, which indicates that optimal orderings which result in changed classification within few actions, aren't likely to be found by chance (with no strong search method). Results for other datasets and models behave similarly. 

Success rates of the baselines are shown in Table \ref{generalised-attack-success}. They show that LUNATC clearly outperforms the simple search attack across all datasets and models, and even outperforms Textfooler's per-sample attack at times. This shows that universal adversarial patterns exist for text data (under a relevant perturbation scheme), and that LUNATC is able to find at least some of them. At a similarity threshold - 0.9, both Genfooler variants perform within the error margin of the simple search baseline, which could indicate that learning the importance as a regression task is over-sensitive to small mistakes, and a rank prediction method is required.

\begin{landscape}

\begin{table}
  \centering
  \begin{tabular}{cclll|ll}
  \toprule
             &                 & \multicolumn{3}{c}{BERT}   & \multicolumn{2}{c}{Word-LSTM}\\
                           & Attack Kind & IMDB  & Toxic & Pubmed & Toxic & Pubmed \\
  \midrule
  \multicolumn{2}{c}{Model Accuracy} & 94.23 & 93.17 & 96.79 & 94.11 & 95.78 \\
  \midrule
  \textbf{per-}   & Textfooler  & 35.97 & 20.33 & 33.40      & 25.8     & 43.5        \\
  \textbf{text}   & PWWS        & 81.19 & 85.17 & 44.04      & 87.95     & 91.16       \\
  \midrule
  \multirow{9}{*}{\rotatebox[origin=c]{90}{\textbf{Universal policy}}} & simple-search & 34.52$\pm$.99    & 30.31$\pm$1.9     & 16.82$\pm$1.0      & 29.5$\pm$1.17     & 26.28$\pm$1.0 \\
  \cmidrule{2-7}
             & gen-PWWS\_500      & 34.98$\pm$3.06 & 29.82$\pm$.98   & -      & 28.31$\pm$.76  & -       \\
             & gen-fooler\_500    & 33.33$\pm$.97  & 28.55$\pm$.69   & 16.49$\pm$.58      & 31.07$\pm$.62  & -       \\
             & LUNATC\textbackslash emb\_500 & 33.99$\pm$1.89 & -     & -      & -     & -       \\
             & LUNATC\_500        & \textbf{39.38$\pm$2.71} & \textbf{36.68$\pm$1.41}  & \textbf{21.81$\pm$.26}  & \textbf{36.03$\pm$.93}     & \textbf{43.59$\pm$.58}       \\
  \cmidrule{2-7}
             & gen-PWWS\_Max      & 34.76$\pm$2.45 & 29.51$\pm$.63 & -      & 28.69$\pm$1.16 & -       \\
             & gen-fooler\_Max    & 34.54$\pm$.31  & 30.38$\pm$.98 & 16.68$\pm$.44      & 30.07$\pm$.79     & -       \\
             & LUNATC\textbackslash emb\_Max & -     & -     & 16.83$\pm$.87      & -     & -       \\
             & LUNATC\_Max        & \textbf{40.04$\pm$.56} & \textbf{37.56$\pm$1.19} & \textbf{29.79$\pm$.79}    & -     & \textbf{45.65$\pm$.52}       \\
    \bottomrule
  \end{tabular}
 \caption{Different attacks' success rate (at .9 similarity threshold) and the classifiers' original test accuracy. Universal attacks state the number of training texts used. \textit{Max} - indicates that all training texts were used, namely 750, 950, 25,000 and 50,000 for IMDB, Toxic-Wikipedia, Pubmed against BERT and pubmed against word-LSTM respectively. \textit{LUNATC\textbackslash emb} is like LUNATC but uses classic DQN and not our method with action embeddings. Due to limited compute resources, we didn't run all attacks on all datasets and models, missing runs are marked ``-"}
 \label{generalised-attack-success}
\end{table}

\end{landscape}

\begin{figure}[t]
  \centering
  \includegraphics[width=\linewidth]{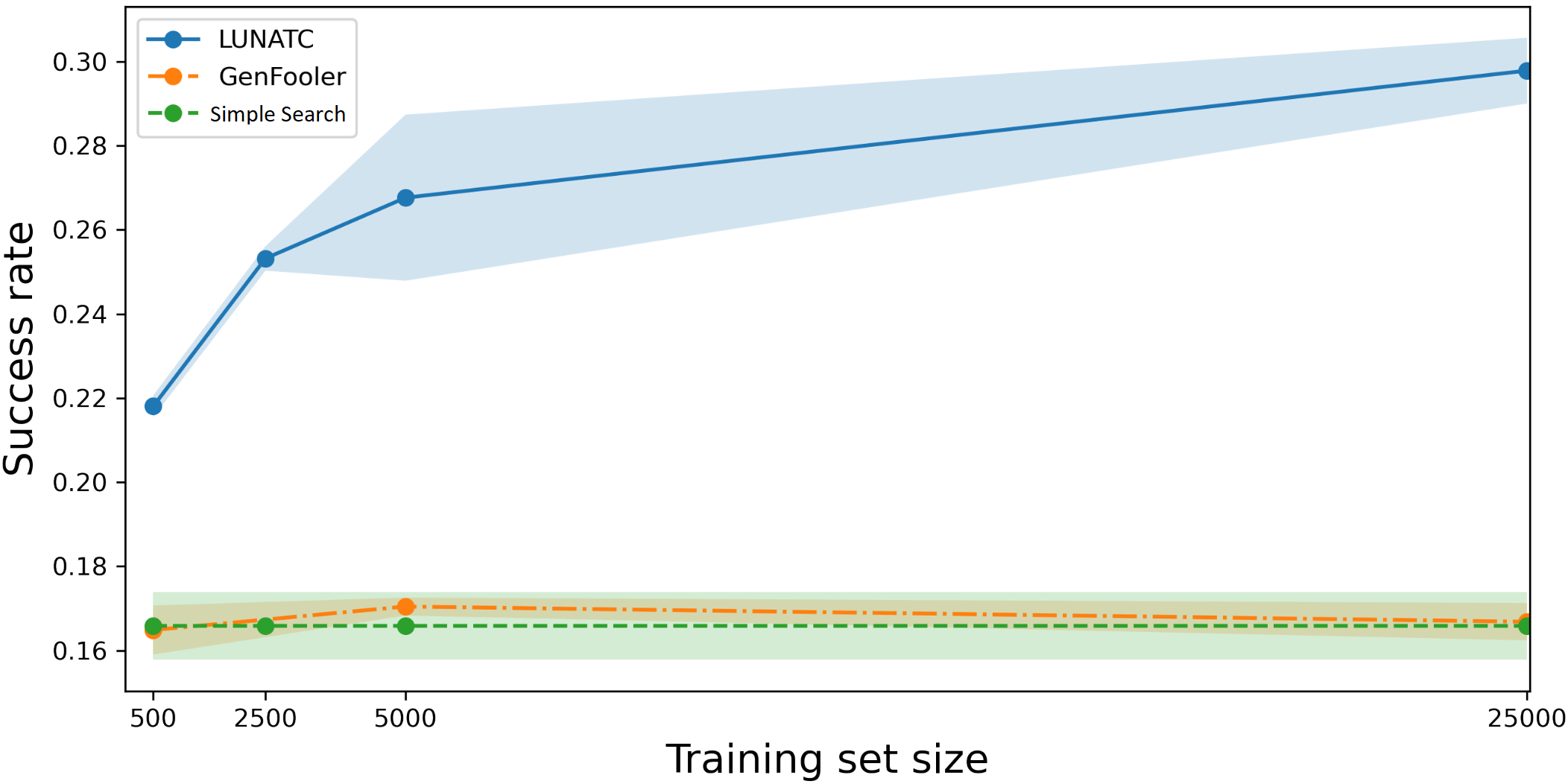}
  \caption{Success rate (at 0.9 similarity threshold) as a function of the number of training texts, on the Pubmed dataset, against a BERT classifier. The error margins indicate the standard deviation across three seeds.}
  \label{fig:sr_train}
\end{figure}

We also wanted to evaluate how adding more training texts would increase generalisation abilities of our approach (and others). As shown in Figure \ref{fig:sr_train} - LUNATC's test performance increases with the training size, and did not reach a plateau in the examined training sizes. It remains for future work to assess if the trend continues for larger training sizes. Conversely, the Genfooler baselines do not seem to improve noticeably with the training size. \textcolor{black}{The error margins are fairly small for LUNATC across all training sizes, especially for the smaller sizes. On further inspection it is also clear that the models succeed on many of the same test texts. This indicates that some texts are ``easier" to attack than others, for instance that many of their possible perturbations have a different predicted class - e.g much of the search space is an adversarial example. Therefore, the lower training sets are successful mainly on these, for which different policies are still likely to succeed. Conversely, LUNATC\_25k is successful on ``hard" examples for which only specific action orderings are able to generate adversarial examples. Which ones succeed and how many varies according to the training process and therefore there is a higher variance, as indicated by the shaded area.}

We also perform an ablation study to assess the impact of the DQN version that uses embeddings to represent actions. The results of the LUNATC attack with the classic DQN version appear in Table \ref{generalised-attack-success}, as \textit{LUNATC\textbackslash emb}. We can see that this approach did not manage to generalise to unseen texts (though as successful as LUNATC on train texts), this indicates that the location of words is not informative enough to estimate their ``importance".

\begin{figure}[t]
  \centering
  \includegraphics[width=\linewidth]{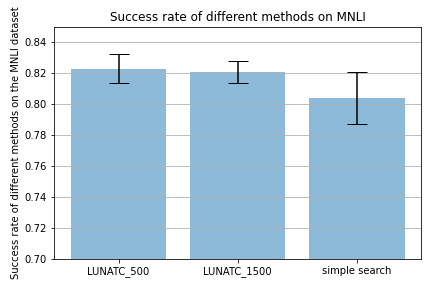}
  \caption{\textcolor{black}{Success rate (at 0.9 similarity threshold) for LUNATC with different training sizes compared to the simple search method, on the MNLI dataset, against a BERT classifier. The error margins indicate the standard deviation across three seeds for LUNATC, and 10 seeds for simple search.}}
  \label{fig:mnli}
\end{figure}

\textcolor{black}{We also demonstrate that our approach also shows generalisation in the task of natural language inference on the MNLI dataset. The results are brought in Figure \ref{fig:mnli}. The general success rates of various attacks, including the simple search method, are higher on these datasets, which we believe indicates the model's sensitivity and brittleness. This in turn means that the relative improvement from learning isn't as big, and yet the mean success rates are clearly higher for the learning approaches compared to the simple search which is non-universal.}

\begin{figure*}[!ht]
  \centering
  \includegraphics[width=\linewidth]{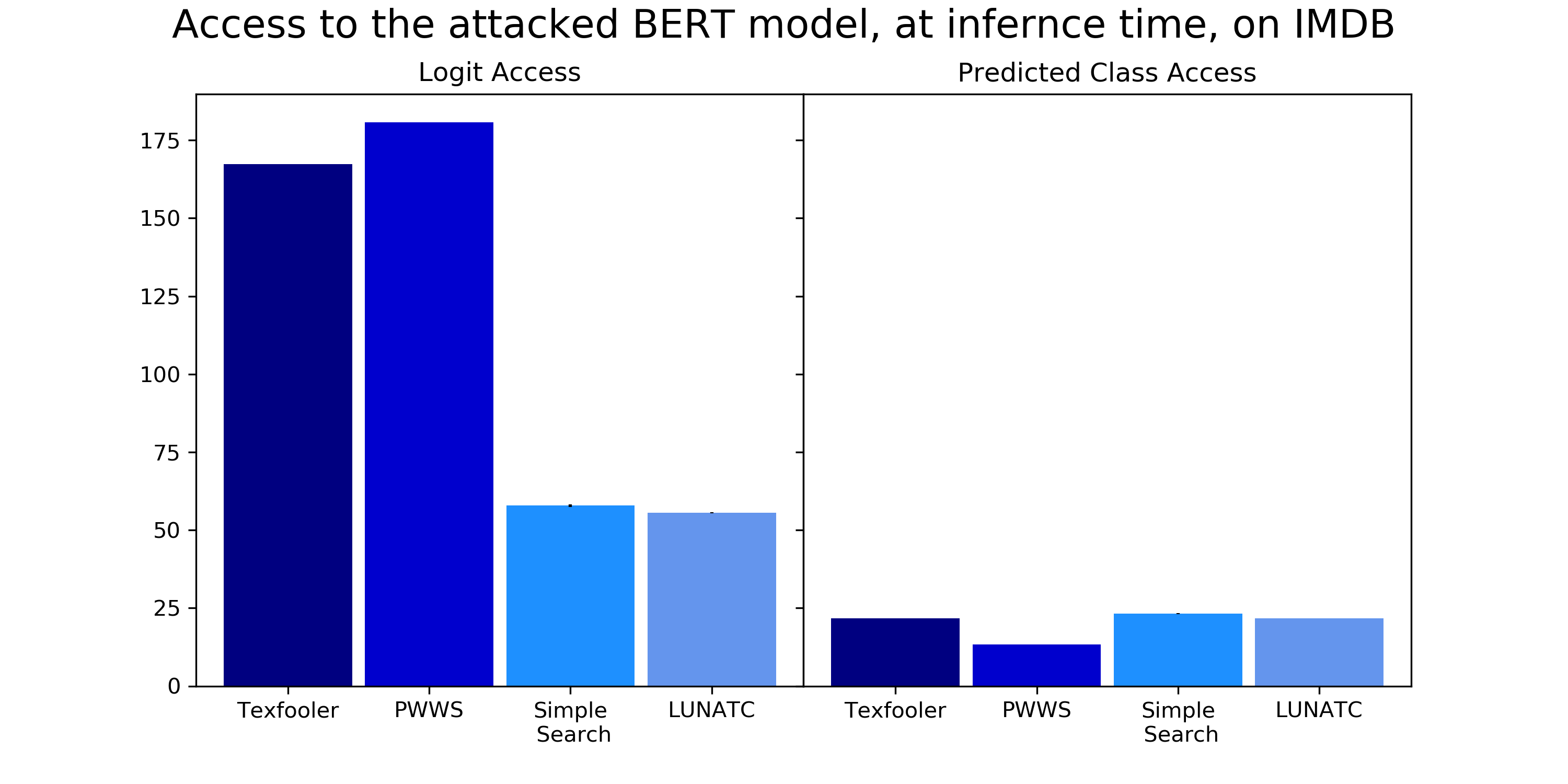}
  \caption{This figure compares the oracle access of different baselines to the attacked BERT classifier at test time (on unseen texts). We report results on the IMDB dataset, though other datasets behaves similarly. We separate the access to predicted class access and logit access. Logit access is harder to get in real life situations and is used for selecting which synonym to use by all methods. Textfooler and PWWS also use it for choosing which word to replace, hence their access is much higher than the LUNATC and simple search. The error margins for the stochastic methods, indicate the mean of three runs for LUNATC, and 10 for the simple search, and indicate a 95 percent confidence interval. However, they are all less than 0.5 thus hardly visible.}
  \label{fig:oracle_access}
\end{figure*}

Finally, we claim our approach is more efficient in test time compared to per-sample attacks, regarding oracle access, because model access is not required for selecting which word to replace (as in per-sample attacks). Instead, the model is used to assert when the class changed - which does not require logit access. In this comparison - the synonym method is greedy and therefore black-box model access is used for synonym selection as well, in the same way for all methods. As discussed in section \ref{sec:conclusion} we hope this need will also be alleviated, in the future thus allowing for no model access at test time. Figure \ref{fig:oracle_access} clearly supports our claim, and shows that our model access is comparable to, and even less than, that of the basic-search, while being notablly less than the per-sample attacks (Textfooler and PWWS).

\section{Conclusions and Discussion}\label{sec:conclusion}
Our results show that universal adversarial policies, as defined in this study, do exist and can generalise to unseen texts from as little as 500 training texts. We further saw how adding more training texts consistently improved the results.

As the results show, our RL-based method - LUNATC, clearly outperformed all baselines for the formulation. We hope this work leads to further research which will improve results by utilising advances in RL or other methods. We hope that future work assesses the agent's ability to use non-greedy actions, and a \textit{stop-game} action, instead of relying on oracle access to check for classification change after each action. These changes will remove the need for test-time oracle access and improve success rates further thus mitigating any compromises compared to trigger based universal attacks further. Analysing what characterises the texts to which the agent successfully generalises also remains for future work.   

LUNATC improved results across all similarity thresholds, however, recent research addressed other validity metrics such as grammatical correctness or ``non-suspicion". They demanded threshold of 0.98 on the similarity to get natural results, though the use of RL could optimise for these directly. By adding a reward term relating to the validity or likelihood of the output text, such as a language model's perplexity, we could learn to generate more natural adversarial texts. This is like approaches performed in other domains \citep{sharif2016accessorize} and also studied in the text domain.

Recent work suggested defence methods against adversarial attacks on text classifiers. \Citet{xu2019lexicalat} suggested a \textit{reinforcement learning}-based adversarial training method which claims to improve model robustness but was not evaluated against strong attack methods. In future work, we wish to evaluate ourselves against ``defended" classifiers, and also see if using our attack for adversarial training can improve robustness. We also wish to add the defence directly to the optimisation task as part of the reward and see if that allows us to break it, akin to \citet{athalye2018obfuscated}.

Finally, we wish to further the study of generalisation in adversarial examples to text classifiers by evaluating the generalisation between models as well. This will hopefully further our understanding of text adversarial examples.

\bibliographystyle{model1-num-names}
\bibliography{nn}

\end{document}